\documentclass{SCIS2022}
\usepackage{cite}
\usepackage{color}
\usepackage[normalem]{ulem} %for deleting lines
\usepackage{amsmath}
\usepackage{multirow}
\usepackage{booktabs}
\usepackage{hyperref}

\begin{document}
%\oa
%%%%%%%%%%%%%%%%%%%%%%%%%%%%%%%%%%%%%%%%%%%%%%%%%%%%%%%
%%% Authors do not modify the information below
\ArticleType{LETTER}
%\SpecialTopic{}
\Year{2023}
\Month{September}
\Vol{66}
\No{9}
\DOI{10.1007/s11432-022-3675-1}
\ArtNo{199101}
\ReceiveDate{8 July 2022}
\ReviseDate{7 October 2022}
\AcceptDate{29 November 2022}
\OnlineDate{1 August 2023}
%%%%%%%%%%%%%%%%%%%%%%%%%%%%%%%%%%%%%%%%%%%%%%%%%%%%%%%

\newcommand{\blu}[1]{\textcolor{blue}{#1}}

\definecolor{Rosolic}{cmyk}{0.00,1.00,0.50,0}
\definecolor{bleudefrance}{rgb}{0.19, 0.55, 0.91}
\definecolor{green}{cmyk}{0.3,0.2,0.95,0.0}
\definecolor{brown}{cmyk}{0.4,0.7,1.0,0.5}
\definecolor{dblue}{cmyk}{1,0.97,0.35,0.0}
\definecolor{Cyan}{cmyk}{1,0,0,0}
\definecolor{Azure}{rgb}{0.0, 0.5, 1.0}

\newcommand{\todo}[1]{\marginpar{\Large {\color{red} $\spadesuit$}} {\bf [#1]}}
\newcommand{\z}[1]{{\color{Rosolic} #1}}
\newcommand{\zc}[1]{{\color{blue} \bf [#1]}}
\newcommand{\zs}[1]{{\color{red} \sout {#1}}}
\newcommand{\xchen}[1]{{\color{Azure} #1}}
\newcommand{\xchenc}[1]{{\color{Azure} \bf [#1]}}
\newcommand{\xchens}[1]{{\color{Azure} \sout {#1}}}
\newcommand{\rewrite}[1]{{\color{red}#1}}
\newcommand{\revised}[1]{{\color{bleudefrance}#1}}
\newcommand{\bbj}[1]{{\color{green} #1}}
\newcommand{\bbjs}[1]{{\color{brown} \sout {#1}}}
\newcommand{\bbjc}[1]{{\color{dblue} \bf [#1]}}
\newcommand{\hb}[1]{{\color{red} {#1}}}
\newcommand{\hbc}[1]{{\color{Cyan} {[HB: #1]}}}

\newcommand{\sysname}{NeuralReshaper}
%%% title:
%%%   \title{title}{title for citation}
\title{\sysname: Single-image Human-body Retouching with Deep Neural Networks}

%%% Corresponding author:
%%%   \author[number]{Full name}{{email@xxx.com}}
%%% General author:
%%%   \author[number]{Full name}{}
\author[1]{Beijia Chen}{}
\author[1]{Yuefan Shen}{}
\author[2]{Hongbo Fu}{}
\author[1]{Xiang Chen}{}
\author[1]{Kun Zhou}{}
\author[1]{Youyi Zheng}{{youyizheng@zju.edu.cn}}

%%% Author information for page head.
\AuthorMark{Beijia Chen}

%%% Authors for citation.
\AuthorCitation{Chen B J, Shen Y F, Fu H B, et al. \sysname: Single-image Human-body Retouching with Deep Neural Networks}

%%% Authors' contribution.
%\contributions{Authors A and B have the same contribution to this work.}

%%% Address.
%%%   \address[number]{Affiliation, City {\rm Postcode}, Country}
\address[1]{State Key Laboratory of CAD\&CG, Zhejiang University, Hangzhou, China}
\address[2]{School of Creative Media, City University of Hong Kong, Hong Kong, China}

\maketitle

%%%%%%%%%%%%%%%%%%%%%%%%%%%%%%%%%%%%%%%%%%%%%%%%%%%%%%%
%%% The main text.
%%%%%%%%%%%%%%%%%%%%%%%%%%%%%%%%%%%%%%%%%%%%%%%%%%%%%%%
\begin{multicols}{2}
Semantic retouching of human bodies in images, such as increasing the height and slimming the body, has been long desired. However, the problem is essentially ill-posed because one should anticipate a set of articulated and nonrigid deformations of different body parts, given that the deformations are inherently three-dimensional. This situation becomes more complicated when images are captured in unrestricted environments with occlusions, and complex interactions between the human body and its surroundings. Early attempts \cite{zhou2010parametric} have attempted to address this issue by interactively fitting a 3D parametric human model to human bodies in images and allowing the fitted 3D model to delegate the transformation via image warping. Although compelling results are produced, these methods may suffer from laborious interactions and foreground and background distortions. % the following two aspects. First, manual efforts are often required to iteratively correct the body-to-image fitting to ensure plausible retouching. Second, warping-based reshaping unavoidably introduces distortions in both the foreground and the background under large deformations.

\begin{figure*}
	\centering
	\includegraphics[width=0.98\linewidth]{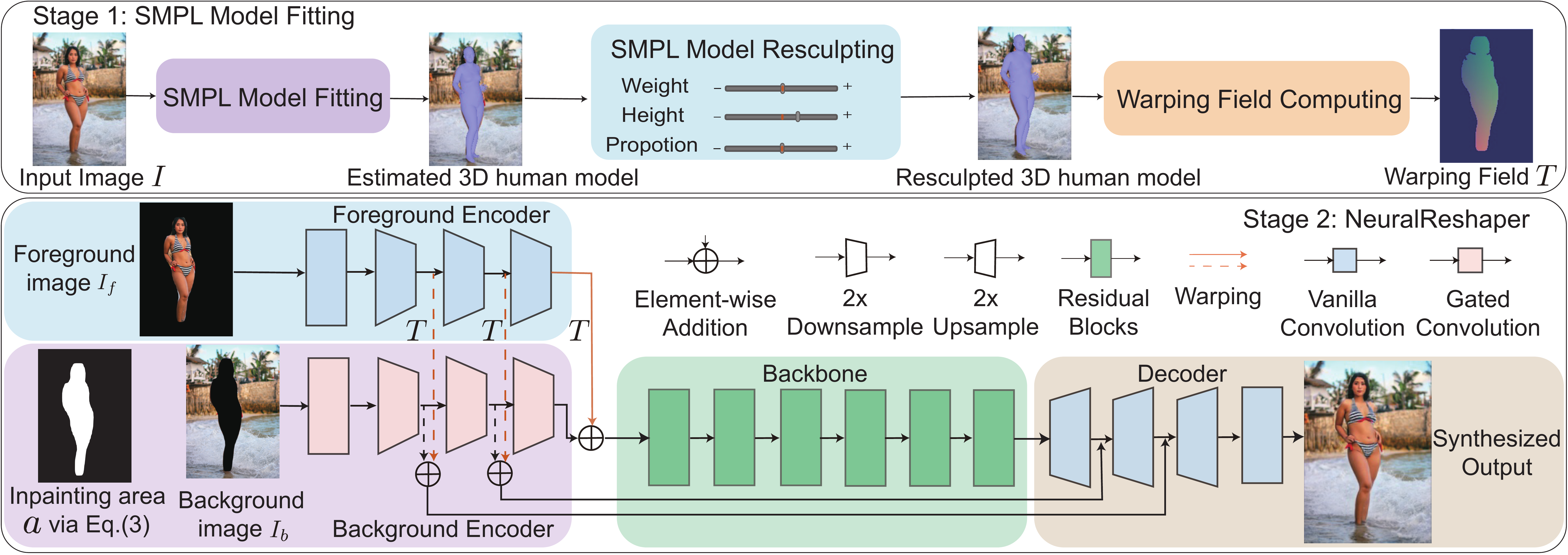}
	\caption{Overall pipeline of our proposed method, consisting of model fitting and neural reshaping stages.}
	\label{fig:pipeline}
\end{figure*}

Inspired by the impressive synthesized images from GANs~\cite{lim2017geometric}, we present \sysname~(Fig.~\ref{fig:pipeline}), the first self-supervised learning-based method for realistic human-body reshaping in a single RGB image following a fit-then-reshape paradigm. The fitting process was first automated  using a hybrid learning-and-optimization-based method with the skinned multiperson linear (SMPL) \cite{loper2015smpl} model.
Then, in an essential stage, the 3D geometric deformation derived from the SMPL model was used to guide the synthesis of the image reshape results. A specific set of design strategies are incorporated into our pipeline to achieve a faithful reshaping result. First, the synthesis process was divided into foreground and background and presented with two independent encoders to ease the reshape learning holistically. Second, to address structure misalignment, the 3D body deformations within our network via feature space warping were incorporated for foreground encodings. The warped foreground encodings are further combined with the background encodings before being passed to a decoder to produce a final reshaped image. Finally, to address the lack-of-paired-data problem, a novel self-supervised strategy was introduced to train our network with our pseudo-paired data. 

\sysname~is fast to use, fully automatic, and robust for images taken in unconstrained environments. The independent nature of SMPL parameters enables us to provide users with high-level semantic control over several key attributes of the human body, such as height and weight. We compare our method to several previous works and possible deep learning baselines. The evaluation of the indoor and in-the-wild datasets shows the superiority of our proposed method over the previous art and alternative solutions.

\textbf{Method.} We \textit{automate} the parametric body fitting process using a data-driven initial fitting followed by a fine-tuning optimization step. For realistic reshaping, we introduce a novel \textit{two-headed} neural architecture.

\textit{Skinned Multiperson Linear Model Fitting. } 
SMPL is a differentiable mapping from the shape $\beta \in \mathbb{R}^{10}$ and pose parameters $\theta \in \mathbb{R}^{72}$ to a 3D human model $M(\beta, \theta)$. Given a human image $I$, we obtain the initial shape and pose parameters $(\beta^{0}, \theta^{0})$ with the pretrained model of \cite{kanazawa2018end}. To further refine the SMPL parameters, we used an optimization-based paradigm to iteratively align the fitted model with respect to the image cues. The overall optimization consists of two steps: optimizing $\beta$ and $\theta$ using 2D key points (following~\cite{kolotouros2019learning}) and optimizing $\beta$ for better silhouette matching. Because of the inherently decoupled shape and pose parameters in SMPL, users can intuitively achieve the desired body shape by directly adjusting the shape parameters $\beta$.

We project the corresponding 3D deformation onto the image space for the resculpted 3D human model to prepare a dense warping field $T$ for the subsequent image reshaping stage. Given the dense warping field $T$, a naive idea would be to directly warp at the pixel level. However, a direct image warping approach based on $T$ and its extrapolation from the human region to the entire image would easily result in noticeable artifacts in the background and foreground. In contrast, we choose to use $T$ to guide the subsequent image synthesis via a neural generator to avoid distortions.

\textit{\sysname. } 
As shown in Fig.~\ref{fig:pipeline}, our network is designed as a two-headed UNet-like structure containing two encoders and a decoder to disentangle the complex foreground--background interactions. The foreground encoder $\mathcal{E}_f(I_f)$  consumes a foreground image $I_f$, the background encoder $\mathcal{E}_b(I_b,a)$ consumes a background image $I_b$ with masked regions and a union foreground mask $a$ (including occluded and disoccluded areas induced by deformation), and the decoder $\mathcal{D}$ takes the encoded codes from $\mathcal{E}_f$ and $\mathcal{E}_b$ and generates the final retouched result $I^{t} $. In an essential step, we take the warping field $T$ derived from the body deformation to warp foreground features and fuse-warped foreground features with encoded background features. In addition to the above generator, during the training stage, a discriminator is used to enforce the overall realism of the generated image.

We introduce a unique warp-guided mechanism to integrate the features of the two encoders to generate desired retouching results based on $T$. Specifically, let $f_{i}$ denote the intermediate feature produced by the $i$-th layer of the foreground encoder $\mathcal{E}_f$, i.e., $f_i = \mathcal{E}_f^i(I_f)$. We warp it to create a distorted feature $f_{i}^{t}=warp(f_{i}, T)$ (shown by the arrows in orange in Fig. \ref{fig:pipeline}), which is roughly aligned with the target shape. Then, we combined the warped foreground feature $f_{i}^{t}$ with the corresponding background feature $b_{i}$ to obtain a complete feature map $\varphi_{i} = f_{i}^{t} + b_{i}$ of the target.

With the warping-aware integration strategy, the generator succeeds in producing a synthesized image $I_{\mathrm{out}} = \mathcal{D}(\mathcal{E}_f(I_f), \mathcal{E}_d(I_b, a), T)$ with the person in the appropriate shape. Ultimately, we obtain the target image
\begin{equation}
	\label{out}
	I^{t} = a \ast I_{\mathrm{out}} + (1-a) \ast I,
\end{equation}
where $\ast$ denotes the element-wise multiplication.

Ideally, we should train our network with paired images $(I, I^{t})$ of the same persons under the same poses but in different shapes. However, obtaining such paired data is difficult, if not impossible. 
For that purpose, we introduce a novel self-supervised training strategy in which we use a deformed source image as the input and attempt to generate the original source image. As a result, the source image can naturally serve as supervisory information without any additional annotation.

To this end, we adopt an L1 loss to encourage this source image recovery
\begin{equation}
L_{R} = \left \| I- G(I_{b}, I_{f}^{t}, T^{t}) \right \|_{1}.
\end{equation}
We used the hinge loss \cite{lim2017geometric} for GAN.
Overall, we obtain an alternating minimization:
\begin{equation}
\begin{aligned}
	\min_{G}\;&(\lambda_{\mathrm{recovery}} L_{R} + \lambda_{\mathrm{gan}} L_{G})\\
	\min_{D}\;&L_{D}
\end{aligned}
\end{equation}
\noindent where $\lambda_{x}$s are tradeoff parameters for different losses.

Appendix C shows more experimental details. Note that our method is desired for altering the human shape parameters such as height and weight. Thus, the human pose is maintained untouched throughout the reshaping.

\textbf{Conclusion. }\sysname is a practical method for the realistic reshaping of the human body in single images using deep generative networks. Our method enables users to reshape human images by moving several sliders and receive immediate feedback. Extensive findings on the indoor and outdoor datasets and online images have shown our method's superiority compared with alternative solutions. Furthermore, we believe that our method can serve as automatic dataset generation for future supervised learning-based methods.

%%%%%%%%%%%%%%%%%%%%%%%%%%%%%%%%%%%%%%%%%%%%%%%%%%%%%%%
%%% Acknowledgements.
%%%%%%%%%%%%%%%%%%%%%%%%%%%%%%%%%%%%%%%%%%%%%%%%%%%%%%%
\Acknowledgements{This work was supported in part by the National Natural Science Foundation China (Grant No. 62172363).}

%%%%%%%%%%%%%%%%%%%%%%%%%%%%%%%%%%%%%%%%%%%%%%%%%%%%%%%
%%% Supplements.
%%%%%%%%%%%%%%%%%%%%%%%%%%%%%%%%%%%%%%%%%%%%%%%%%%%%%%%
\Supplements{
\href{https://link.springer.com/article/10.1007/s11432-022-3675-1}{Appendix A-C.}
}

\bibliographystyle{myScis}
\bibliography{paper_arxiv}

\end{multicols}

\newpage

\begin{appendix}

\section{Method Details}

\subsection{SMPL Fitting Optimization}\label{Sec:fitting_stage}

\begin{figure}[h]
	\centering
	\includegraphics[width=0.8\linewidth]{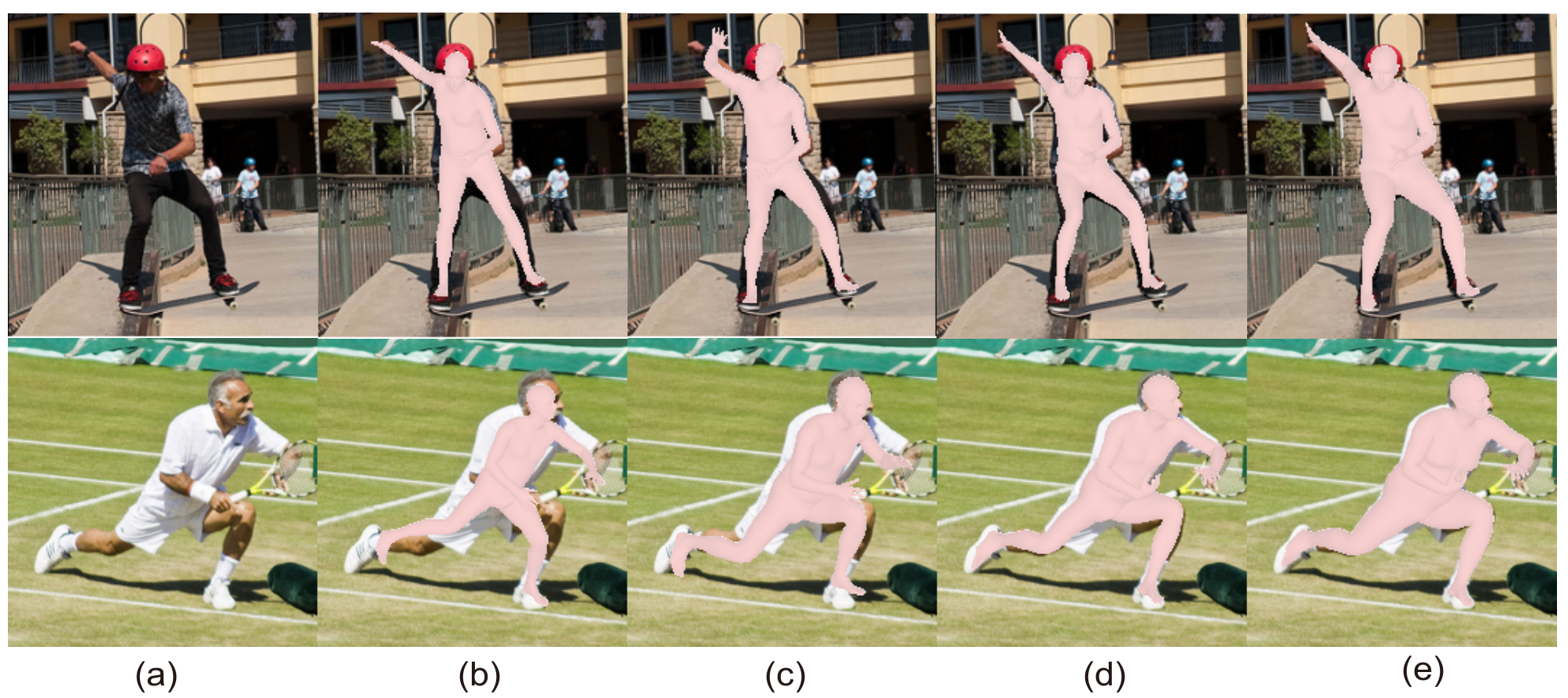}
	\caption[width=10]{Fitting results obtained from different steps in the SMPL model fitting stage. (a) The original image. (b) The optimization result obtained from \cite{bogo2016keep} (initialized with mean parameters). (c) The direct inference result obtained from the pre-trained model \cite{kanazawa2018end}. (d) The fitting result obtained after the 2D keypoints optimization step. (e) The fitting result obtained after the silhouette optimization step.}
	\label{fig:fit_result}
\end{figure}

\textbf{Refining $\beta$ and $\theta$ using 2D Keypoints. } We first extract from $I$ the 2D keypoints $J_{\mathrm{est}} \in \mathbb{R}^{2K}$ along with their confidence $w \in \mathbb{R}^{K}$ by using OpenPose \cite{cao2019openpose} and optimize $\beta$ and $\theta$ to match the image projections of 3D joints with the estimated 2D keypoints $J_{\mathrm{est}}$. Since the initial value of $\beta$ and $\theta$ are also reliable, we adopt a simplified energy of \cite{bogo2016keep} to reduce the optimization time while not harming the accuracy.
\begin{equation}
	E_{\mathrm{joint}}(\beta, \theta)=\sum_{\mathrm{joint}\,i} w_{i} \rho\left(\Pi_{\alpha}M_{\mathrm{joint}}(\beta, \theta)-J_{\mathrm{est}}^i\right),
\end{equation}
where $M_{\mathrm{joint}}$ denotes the 3D SMPL joint positions, $\Pi_{\alpha}$ denotes the orthogonal camera projection and $\rho$ is the robust Geman-McClure penalty function.

\textbf{Refining $\beta$ using 2D Silhouette.}
We perform an additional step to further optimize $\beta$ to minimize the L2 loss between the projected silhouette of the body and the estimated 2D binary mask $m$ of the human body (extracted from $I$ using Mask R-CNN \cite{he2017mask}). We keep $\theta$ intact at this step since optimizing $\theta$ w.r.t. the silhouette may affect the actual positions w.r.t. the 2D key points.

\begin{equation}
	E_{\mathrm{silhouette}}(\beta)=\left \| \mathrm{NR}_{\alpha}[M_{\mathrm{mesh}}(\beta, \theta)] - m \right \|_2^2,
\end{equation}
where $M_{\mathrm{mesh}}$ denotes the SMPL body mesh, and the operator $\mathrm{NR}_{\alpha}[\cdot]$ denotes the differentiable silhouette rendering \cite{kato2018neural} using the camera parameters $\alpha$.

Fig. \ref{fig:fit_result} shows an example with fitting results in different phases. We observe considerable improvements by using both 2D keypoints and silhouette alignment compared to the direct inference results of \cite{kanazawa2018end} (Fig. \ref{fig:fit_result} (b)) and \cite{bogo2016keep} (Fig. \ref{fig:fit_result} (c)), respectively. Note here we do not use the parametric model of SCAPE \cite{anguelov2005scape}, which is used in \cite{zhou2010parametric}, since the SMPL model is more accurate and compatible with modern rendering pipelines \cite{loper2015smpl}.

\begin{figure}[h]
	\centering
	\includegraphics[width=0.55\linewidth]{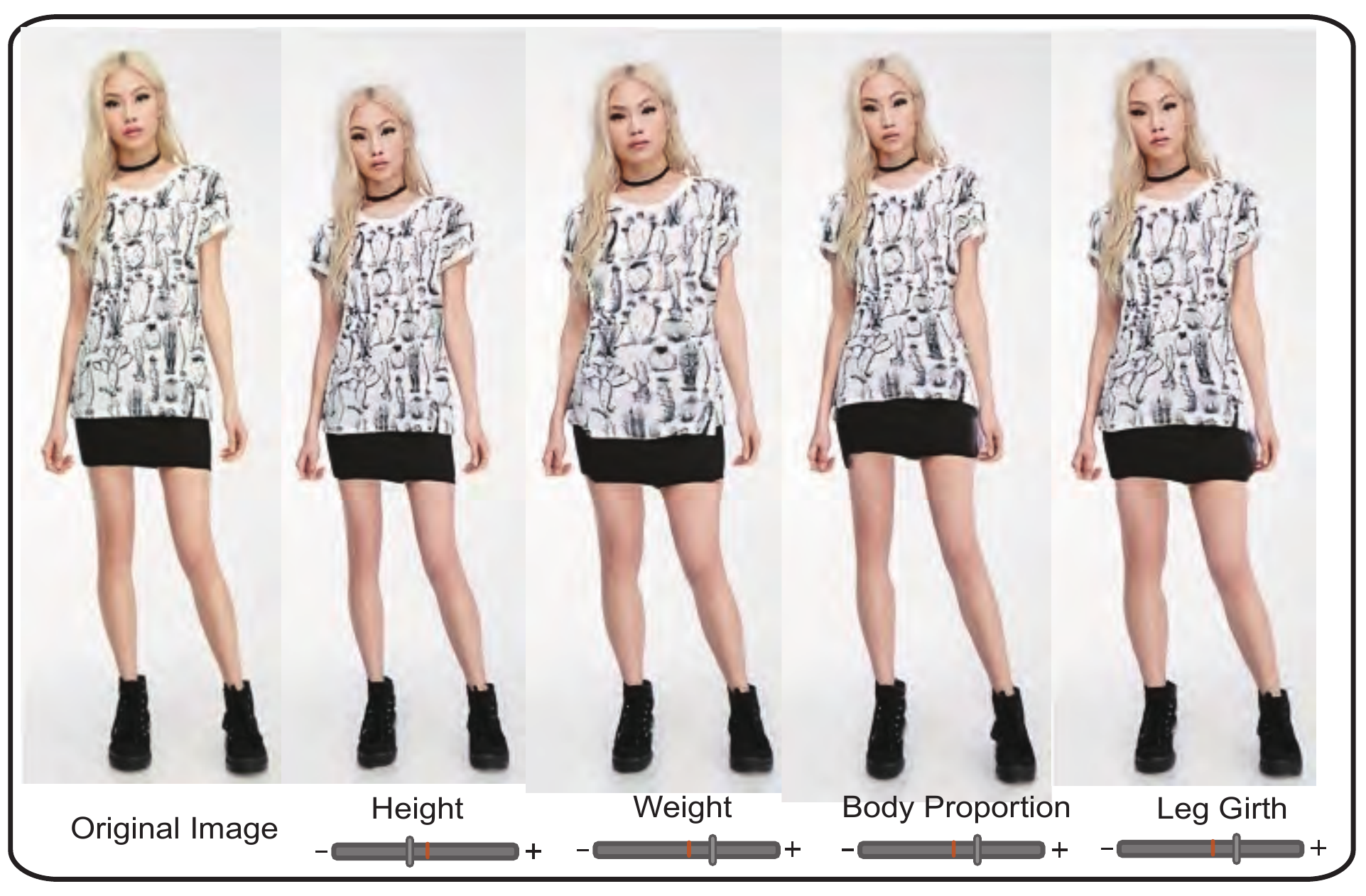}
	\caption[width=18]{The reshaping results generated by our method with attributes adjusted individually.}
	\label{fig:percomp}
\end{figure}

We create an easy-to-use interface for semantic reshaping of human bodies by offering users a set of sliders corresponding to the aforementioned semantic attributes (Fig. \ref{fig:percomp}), and the users can get reshaping results instantly after adjusting. Specifically, we pick four key shape parameters which separately represent the height, weight, leg girth, and body proportion for simple usage. Note that, increasing the body proportion indicates the lengthening of legs while keeping the height unchanged, and vice versa. We also allow users to pick a particular subject by drawing a bounding box before the 3D fitting, if multiple people exist in a source image.

\subsection{Self-Supervised Training Strategy}

\begin{figure*}[t]
	\centering
	\includegraphics[width=\linewidth]{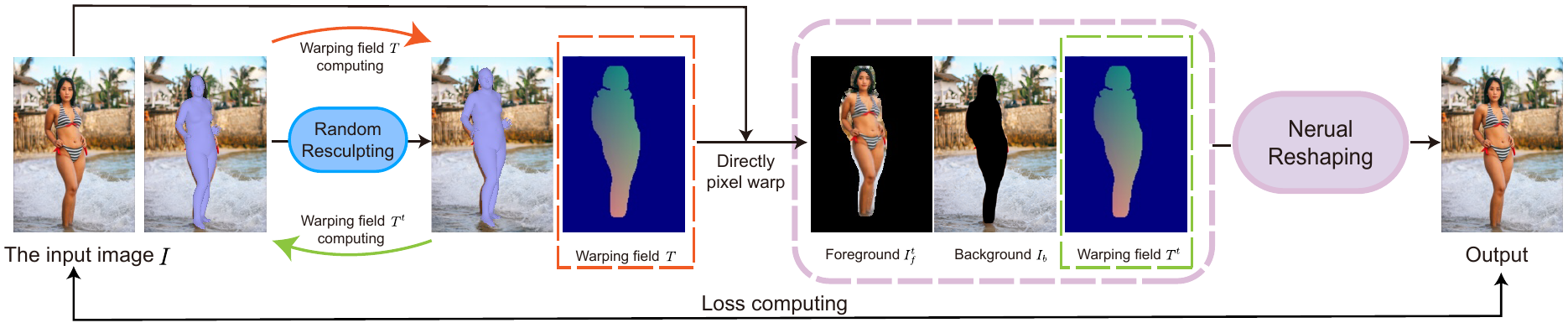}
	\caption{Our self-supervised training strategy. Given an input image $I$ and its fitted SMPL model counterpart, we randomly resculpt the SMPL model and directly warp pixels with the warping field $T$ to get a triplet of pseudo training inputs consisting of the deformed foreground image $I_f^t$, the background image $I_b$ and the inverse warping field $T^t$. The neural reshaping module is trained to recover the original image $I$ from the triplet inputs in which case $I$ provides a neutral supervision.}
	\label{fig:selfsupervised}
\end{figure*}

Specifically, as shown in Fig. \ref{fig:selfsupervised}, for each image from an existing dataset (we denote these images as source images in the following), we first fit SMPL to it as described in Sec. \ref{Sec:fitting_stage}. We then resculpt the fitted 3D model by randomly altering the shape parameters. To ensure the deformation plausibility and generating large deformations simultaneously, we define valid ranges for SMPL shape parameters. Next, we randomly sample shape variations from these valid ranges following the unit Gaussian distribution. All the shape variations are controlled to weight/height changes in the range of [0-20] kg/cm, which are consistent with previous work \cite{zhou2010parametric} and satisfy most of the real-world reshaping requirements. The deformation from the source shape to the deformed one is projected onto the image space to form a warping field, denoted as $T$. Next, we directly warp the source image by $T$ (at pixel-level) to generate the deformed foreground $I_{f}^{t}$, and compute the background image $I_{b}$. Finally, we obtain a paired data $((I_{b}, I_{f}^{t}, T^{t}), I)$ for training where $T^t$ is the inverse of $T$. By taking the tuple $(I_{b}, I_{f}^{t}, T^{t})$ as input, our \sysname~generator $G$ should produce a complete image as close to the source image $I$ as possible.

We benefit from this training strategy in a significant way. With this strategy, we have sufficient data for training. We keep a neat network and concise loss functions. The network learns how to produce the warped foreground and inpaint the background. It also learns how to composite the two parts together naturally. 

Note that directly warping on pixel-level usually brings significant distortions (shown as $I_{f}^{t}$ in Fig. \ref{fig:selfsupervised}), which makes our training foreground different from our testing foreground. Therefore, to relieve this ambiguity, we choose to warp on high-level features instead of directly warping on pixel-level and use GAN loss for realistic image generation. See Fig. \ref{fig:comparasion_warp} for comparisons between our approach and an alternative solution by directly warping in the image space (instead of the feature space).

\begin{figure*}[h]
	\centering
	\includegraphics[width=\linewidth]{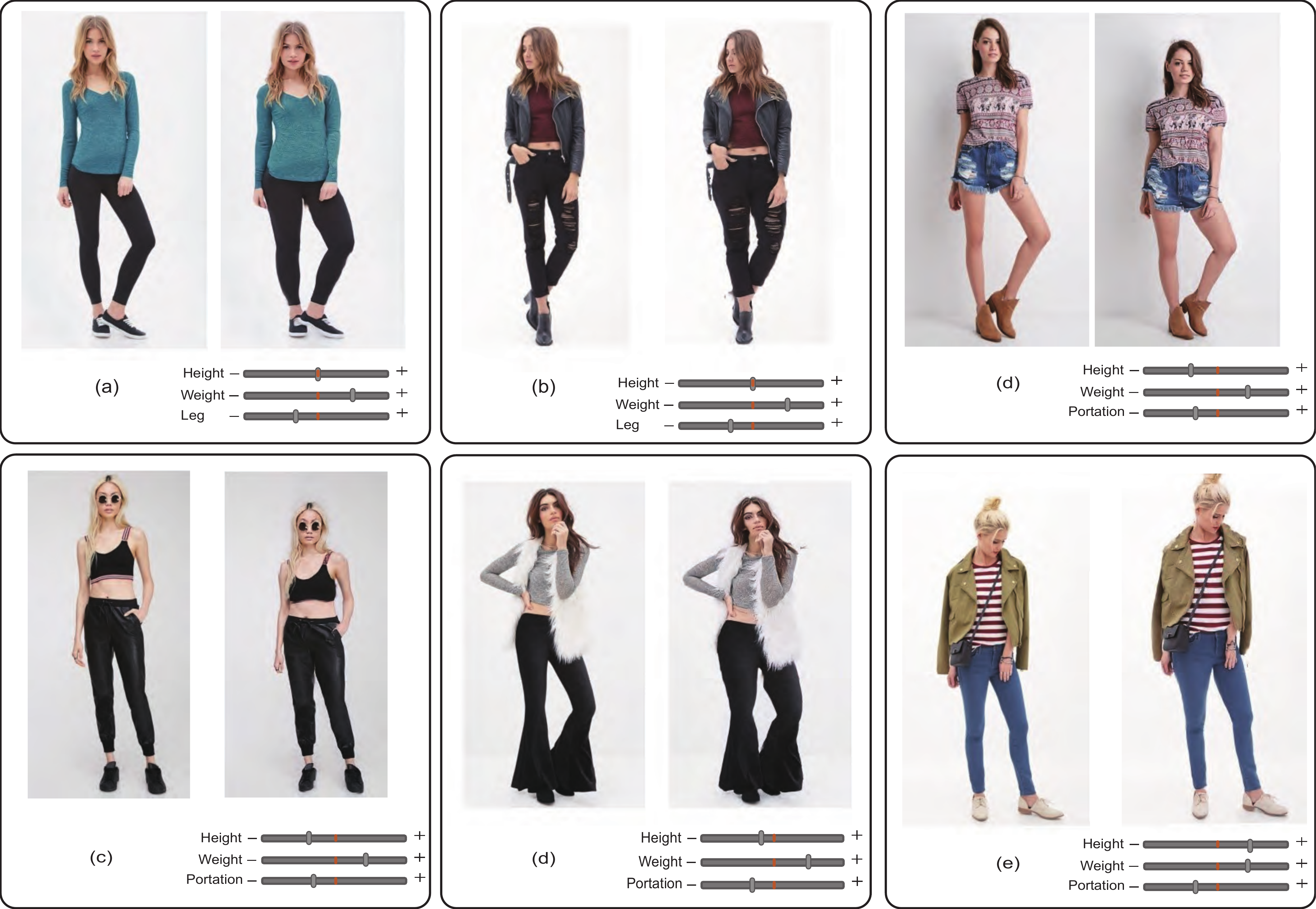}
	\caption[width=15]{A series of representative reshaping results on the DeepFashion dataset. For each case, the left side is the original image and the right side is a reshaped image with the sliders below each case indicating the attributes and their degrees that have been edited.}
	\label{fig:deepfashion_results}
\end{figure*}

\section{Experiments}

\subsection{Dataset}
We have conducted extensive experiments on an indoor dataset DeepFashion \cite{liuLQWTcvpr16DeepFashion} ($512\times512$ resolution) and an outdoor dataset consisting of images from COCO \cite{lin2014microsoft}, MPII \cite{andriluka14cvpr}, and LSP\cite{Johnson10} ($256\times256$ resolution). Since there are fewer high-quality human images in the outdoor dataset, we demonstrate our network's ability to generate photo-realistic images of higher resolution on DeepFashion. Both the training images and testing images are coming from the same source dataset. For DeepFashion, we use 85\% of them for training and the rest for testing. For the outdoor dataset, we collected $29,353$ images in total. Specifically, we randomly pick $24,806$ for training and the rest for testing.

As mentioned in Sec. \ref{Sec:fitting_stage}, we run OpenPose \cite{cao2019openpose} and Mask R-CNN \cite{he2017mask} to obtain the 2D keypoints and human body silhouette for all images. For simplicity, we discard images that have less than six visible keypoints. We fit the SMPL model to each image and get rid of the images if the fitted region does not cover half of the human mask. For each image in DeepFashion, we compute the bounding box of the human and use it to crop the image and resize it to $512 \times 512$. For the outdoor dataset, images are cropped and resized to $256 \times 256$. Though our method proposes to refine SMPL shape parameter according to human body mask, it is still difficult and error-prone to achieve desirable fitting results for loose-fitting cases since SMPL is a naked body model. Thus, we do not consider the case of very loose dresses or skirts in the following experiments.

\subsection{Implementation Details}
\subsubsection{Fitting Stage}
The initial SMPL fitting network \cite{kanazawa2018end, kolotouros2019learning} cannot handle images of varying input size, we crop and resize all training images into $224\times 224$. For the optimization step, we use Adam \cite{kingma2014adam} for both 2D keypoints and silhouette optimization with a learning rate of 0.01 and 0.05, respectively. We set 100 as the total iteration number for both steps. 

\begin{figure*}[t!]
	\centering
	\includegraphics[width=\linewidth]{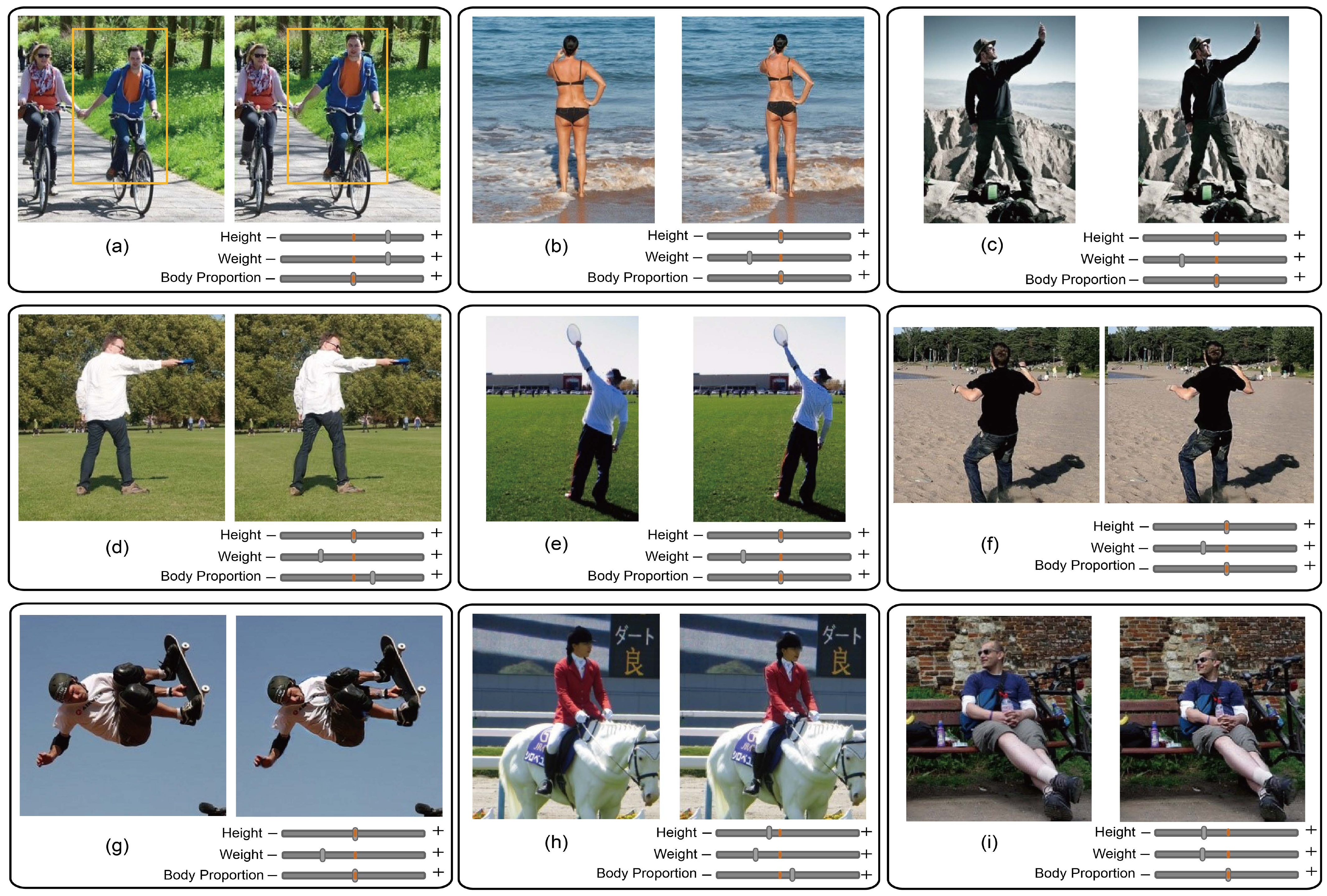}
	\vspace{-0.3cm}
	\caption[width=18]{A series of representative reshaping results for outdoor images. For each case, the left side is the original image and the right side is a reshaped result with the sliders below each case indicating the attributes and their degrees that have been edited. The person that has been reshaped is highlighted with a yellow bounding box in (a).}
	\label{fig:in_the_wild}
	\vspace{-0.3cm}
\end{figure*}

\begin{figure}[t!]
	\centering
	\includegraphics[width=\linewidth]{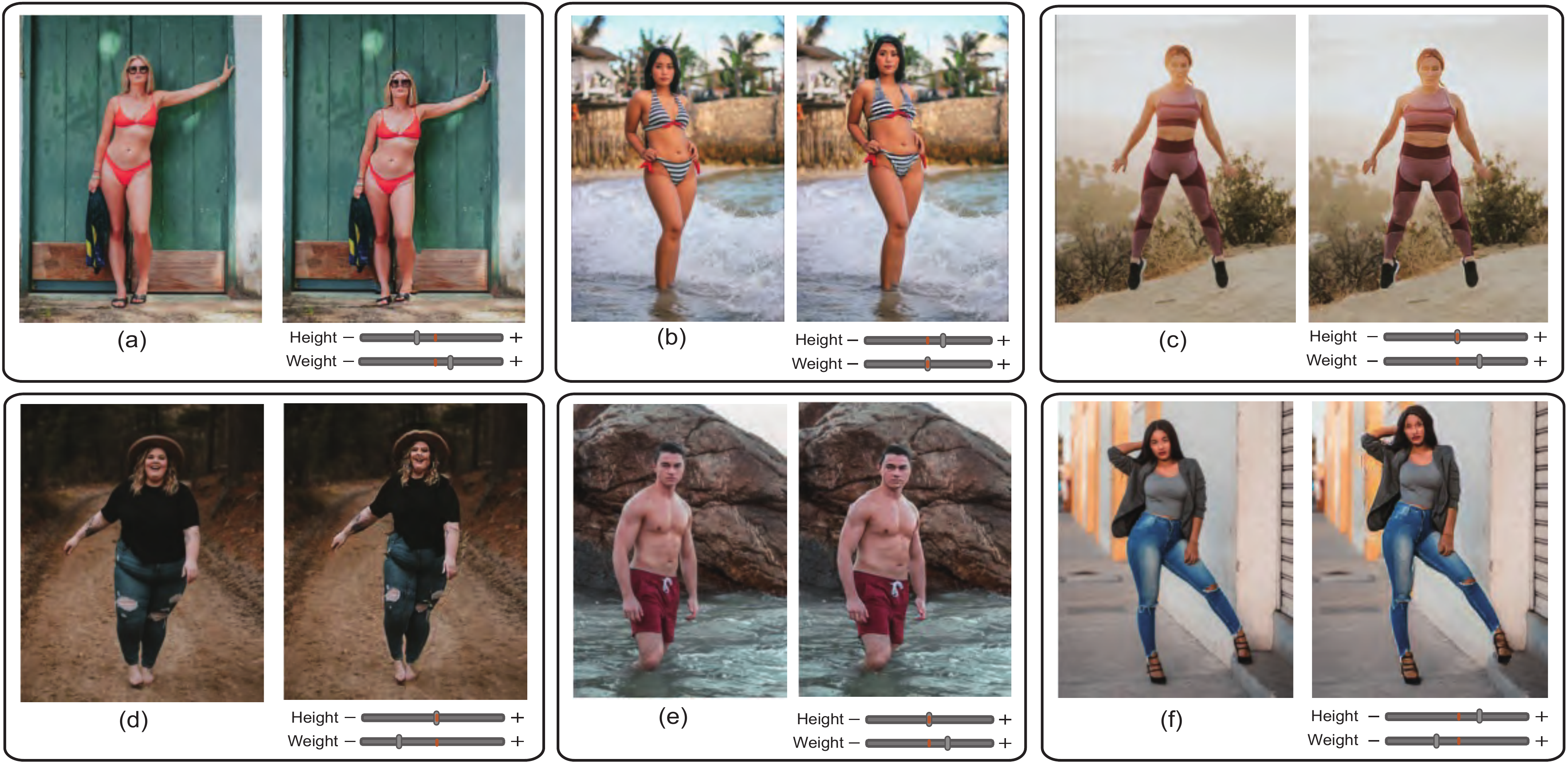}
	\vspace{-0.5cm}
	\caption[width=18]{Several reshaping results tested on online images. For each case, the left side is the original image and the right side is a reshaped result with the sliders indicating the attributes and their degrees that have been edited.}
	\label{fig:internet_cases}
	\vspace{-0.2cm}
\end{figure}

\subsubsection{Reshaping Stage}
{The UNet-like generator consists of two parallel encoders, a bottleneck, and a decoder symmetric to the encoders. Specifically, we employ $4$ layers of convolutions for each encoder. The input layer increases the number of channels to $64$, and the following three downsampling layers decrease the spatial dimension by a factor of $2$ while increasing the numbers of channels to 128/256/512. Each convolution layer is followed by an instance normalization and a leaky ReLU activation. The only difference between the two encoders is that we use gated convolution for the background and vanilla convolution for the foreground since gated convolution provides more flexible feature learning for background inpainting \cite{yu2018generative}. We use $6$ layers of residual blocks in the bottleneck and $4$ layers for the decoder. The spectral-normalized \cite{miyato2018spectral} patch discriminator consists of 6 layers of vanilla convolution.}

We implement our system in PyTorch with one NVIDIA 1080Ti GPU (11 GB memory). We train the network for $100$ epochs with the Adam optimization \cite{kingma2014adam}. We set the learning rates for the generator and discriminator to 0.0001 and 0.0004, respectively. We use a batch size of $8$ for the outdoor dataset and $2$ for the DeepFashion dataset. We set the weight $\lambda_{\mathrm{recovery}}$ to $100$ and $\lambda_{\mathrm{gan}}$ to 10. In the training stage, our method takes about 48 hours for training on the indoor dataset DeepFashion, 72 hours for training on the outdoor dataset. At runtime, our system takes about 15s for pre-processing each image, including semantic segmentation using MaskRCNN~\cite{he2017mask}, keypoint detection by OpenPose~\cite{cao2019openpose}, and SMPL fitting optimization. Then the modification of the parameters to derive reshaping results runs at an instant rate (less than 1s).

\subsection{Qualitative Results}

\subsubsection{Results on the DeepFashion Dataset}
Fig. \ref{fig:deepfashion_results} shows reshaping exemplars on the DeepFashion Dataset with persons in diverse poses, shapes, and appearances. The sliders below each case indicate the user-specified shape attributes. The system supports individual or joint manipulation of these attributes. We observe that the system produces natural and consistent reshaping results while being robust under large deformations. The synthesis results preserve high-frequency details of clothes, face, and hair in the original images. Fig. \ref{fig:percomp} illustrates the individual effect of each attribute, showing the well-disentangled reshaping semantics.

\subsubsection{Results on the Outdoor Dataset}
We also show results on outdoor dataset where complex occlusions and backgrounds can present. Fig. \ref{fig:in_the_wild} shows reshaping exemplars under diverse outdoor situations.
Though trained with the self-supervised data, our system is able to handle large reshaping effects. The synthesized textures in the dis-occluded areas and the deformed foreground are visually realistic. The color discrepancy along the boundary between the mask region and the background is the common artifacts in previous works of image inpainting \cite{yu2019free}. In contrast, our method produces spatially consistent textures and colors in these regions. The synthesized foreground is blended naturally and seamlessly with the background without boundary artifacts. It indicates that our model succeeds in decoding the desired foreground and background in a reasonable spatial layout by merging both the branches' information. Fig. \ref{fig:in_the_wild} shows that our approach provides fine-grained and holistic reshaping effects. %\hbc{Not always: the bike handle in (a) is missing; the left arm in (b) suffers from some artifacts; similar for (c); some details in the right hand (g) are also lost? In (h) there is some artifact in the face region. The fingers in (i) look messy}.
The last row of Fig. \ref{fig:in_the_wild} exemplifies reshaped body images with challenging poses, severe occlusions, and complex backgrounds.

\subsubsection{Results on the Online Images}
Results, exhibited in Fig. \ref{fig:internet_cases}, show that our semantic reshaping method generalizes well to the randomly retrieved online images outside of the testing dataset. It is noteworthy that our method is fully automatic and does not require any annotations such as 2D keypoints or silhouette during testing, thanks to the robust 2D detection techniques (i.e., OpenPose \cite{cao2019openpose} and Mask R-CNN \cite{he2017mask}). Our method is applicable to a wide range of real situations.

\begin{figure}[h]
	\centering
	\includegraphics[width=\linewidth]{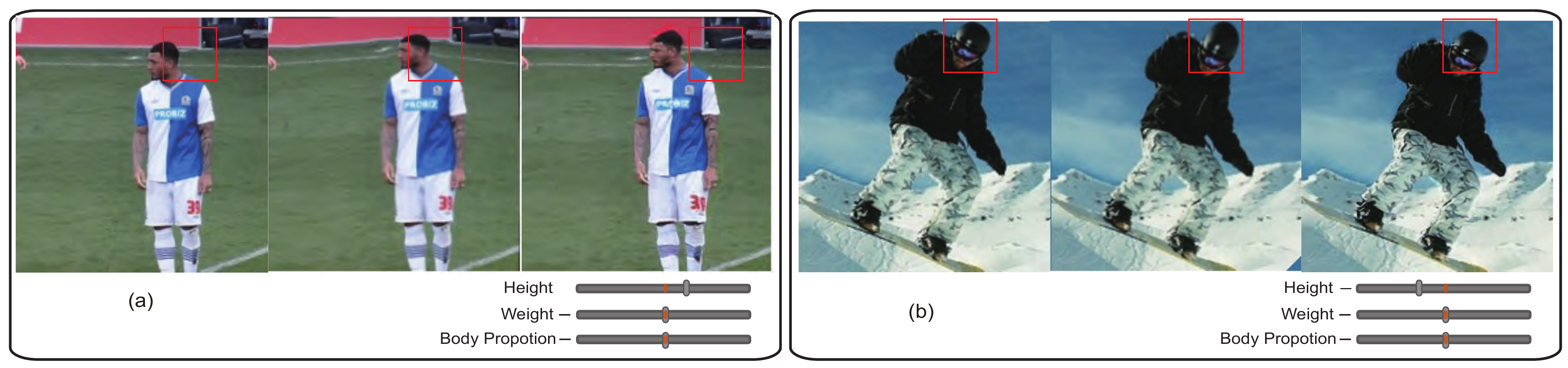}
	\caption{% A comparison with direct warping. % results. 
		A comparison between direct warping with extrapolation and our method.
		From left to right: the original image,
		the reshaping result by direct warping, and the reshaping result by our method.}
	\label{fig:directwarping}
\end{figure}

\subsection{Comparisons}
{In this section, we compare our approach to the direct warping method, the warping-based human reshaping method \cite{zhou2010parametric}, the state-of-the-art human image editing method \cite{liu2019liquid} using deep learning and the state-of-the-art human reshaping method~\cite{ren2022structure}}.

\begin{figure*}[t]
    \centering
    \includegraphics[width=\linewidth]{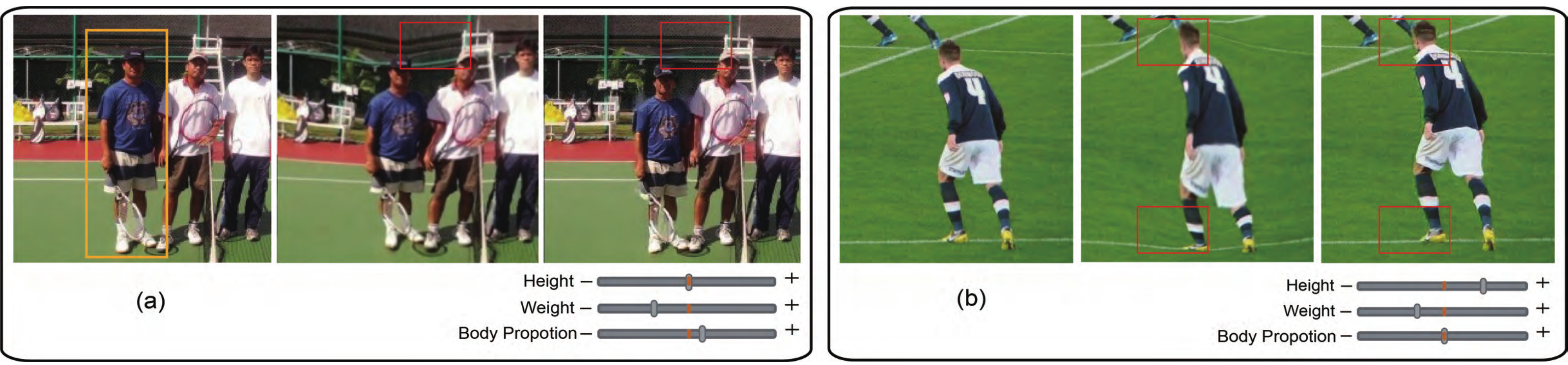}
    \caption[width=18]{Comparison with Zhou et al. \cite{zhou2010parametric}. For each case, from left to right are the original image, the reshaping result by the method in \cite{zhou2010parametric}, and the reshaping result by our method. The sliders below each case indicate the attributes and their degrees that have been edited. The person that has been reshaped is highlighted with a yellow bounding box in (a). We also highlight the distorted areas in \cite{zhou2010parametric} and their corresponding patches in our result with red bounding boxes. }
    \label{fig:comparasion_Zhou}
\end{figure*}

\begin{figure*}[t]
	\centering
	\includegraphics[width=0.8\linewidth]{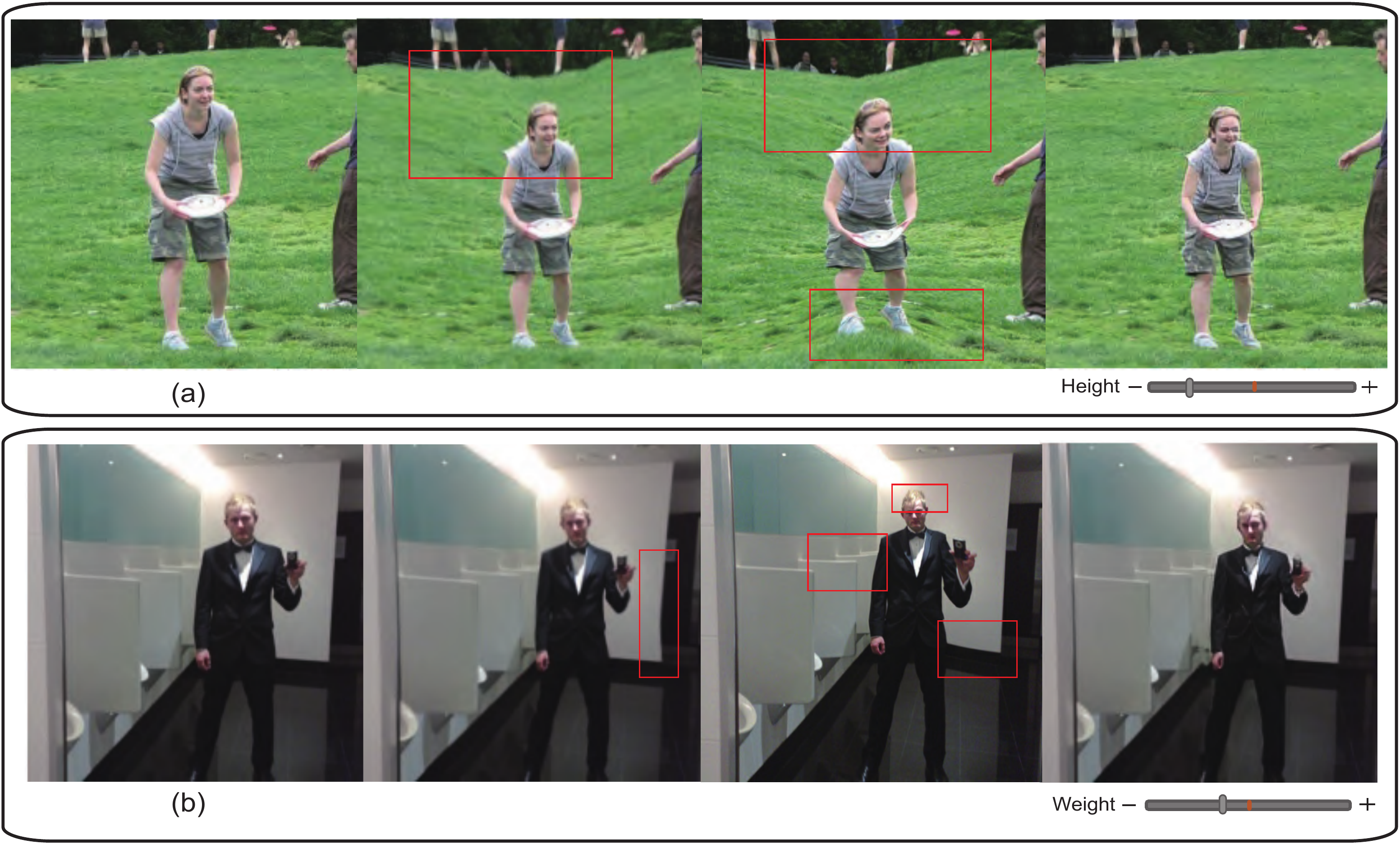}
	\caption[width=18]{Comparisons with the direct warping method and Zhou et al. \cite{zhou2010parametric}. For each case, from left to right are the original image and the reshaping results generated by direct warping, Zhou et al. \cite{zhou2010parametric}, and our method respectively. The sliders below each case indicate the attributes and their degrees that have been edited. We also highlight the distorted areas with red bounding boxes. }
	\label{fig:comparasion_distortion}
\end{figure*}

Visual comparisons with the direct warping method and the warping-based human reshaping method are shown in Fig. \ref{fig:directwarping}, Fig. \ref{fig:comparasion_distortion} and Fig. \ref{fig:comparasion_Zhou}. As one can see, both directly warping-based methods introduce distortions in the foreground and background, leading to unrealistic and implausible results. The larger deformations conducted, the severe distortions induced. This is mainly because that directly warping-based methods first conduct delaunay triangulation on the full image and deform these triangles according to the 3d human model. Thus, the deformations of human model are propagated to the whole image, introducing unfavorable changes on background. Despite the background distortions, warping-based methods also induce distortions on unfitted body areas such as hair (see Fig. \ref{fig:comparasion_distortion} (a)). This is mainly because that the unfitted body areas are treated as background. As suggested in~\cite{zhou2010parametric}, one possible strategy to alleviate such distortion artifacts is to manually adjust the saliency map. Nevertheless, this involves lots of human intervention. In contrast, our method is fully automatic in fitting, and preserves regular patterns occluded by humans (e.g., wire fence in Fig. \ref{fig:comparasion_Zhou} (a)) rather than distorting them. This is because that our method decomposes the generation of foreground and background parts and the generator learns to inpaint the missing areas.

\begin{figure*}[t]
	\centering
	\includegraphics[width=\linewidth]{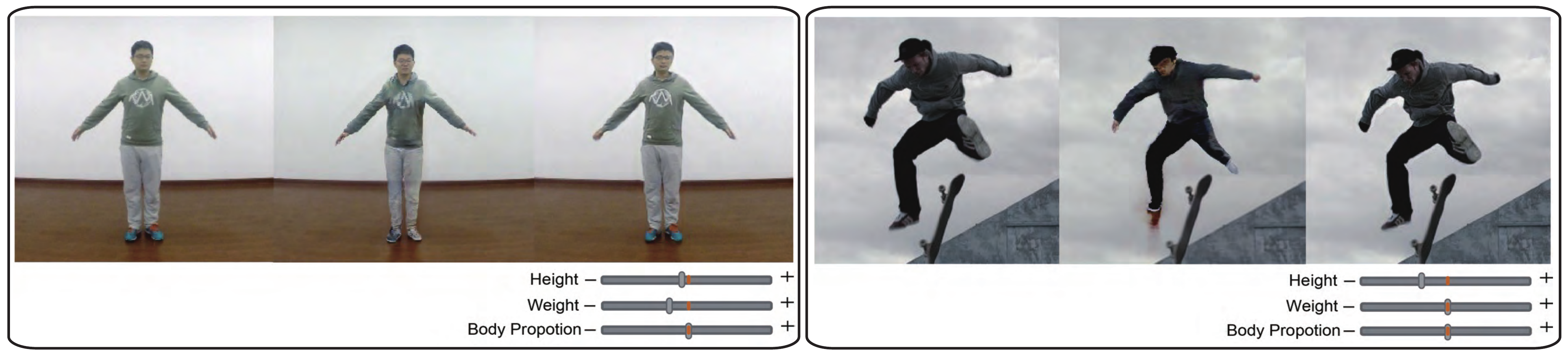}
	\caption[width=18]{Comparison with Liu et al. \cite{liu2019liquid}. For each case, from left to right are the original image, the reshaping result by the method in \cite{liu2019liquid}, and the reshaping result by our method.} 
	\label{fig:comparasion_liquid}
\end{figure*}

The human image editing method \cite{liu2019liquid} trains their network with paired images sampled from videos in a fully supervised manner and applies it to specific tasks like pose-guided human image transfer. For our body reshaping task, since we lack paired data, it is impossible to reproduce their method on outdoor in-the-wild datasets like COCO~\cite{lin2014microsoft}. We compare our method with~\cite{liu2019liquid} on their dataset, COCO dataset and online images, respectively. Fig. \ref{fig:comparasion_liquid} (a) shows that our self-supervised method even gets better results than their supervised training method on the reshaping task. One can see that our method can preserve original appearances well and our results look more realistic. As shown in Fig. \ref{fig:comparasion_liquid} (b), their method leads to seriously degraded results when facing non-seen cases. We also do quantitative comparison with~\cite{liu2019liquid} in Table. \ref{tab:fid}.  We show the Frechet Inception Distance (FID) between randomly sampled generated images and original images. The superiority proves that our method can generate more realistic images again.

\begin{figure*}[h]
	\centering
	\includegraphics[width=0.9\linewidth]{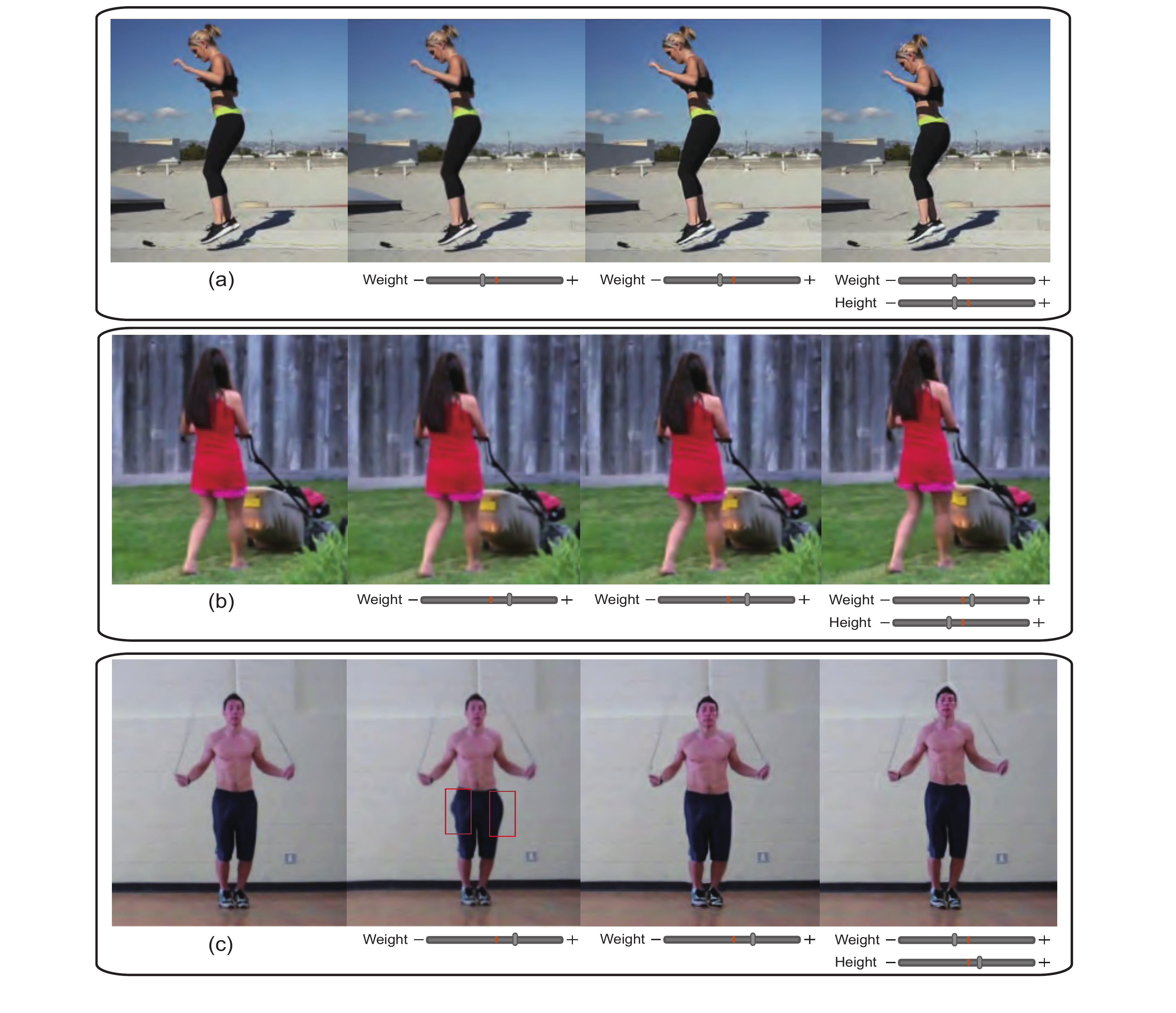}
	\caption[width=18]{Comparison with Ren et al. \cite{ren2022structure}. For each case, from left to right are the original image, the reshaping result by \cite{ren2022structure}, and two different reshaping results by our method.} 
	\label{fig:comparasion_sota}
\end{figure*}

\begin{figure*}[t!]
	\centering
	\includegraphics[width=\linewidth]     
        {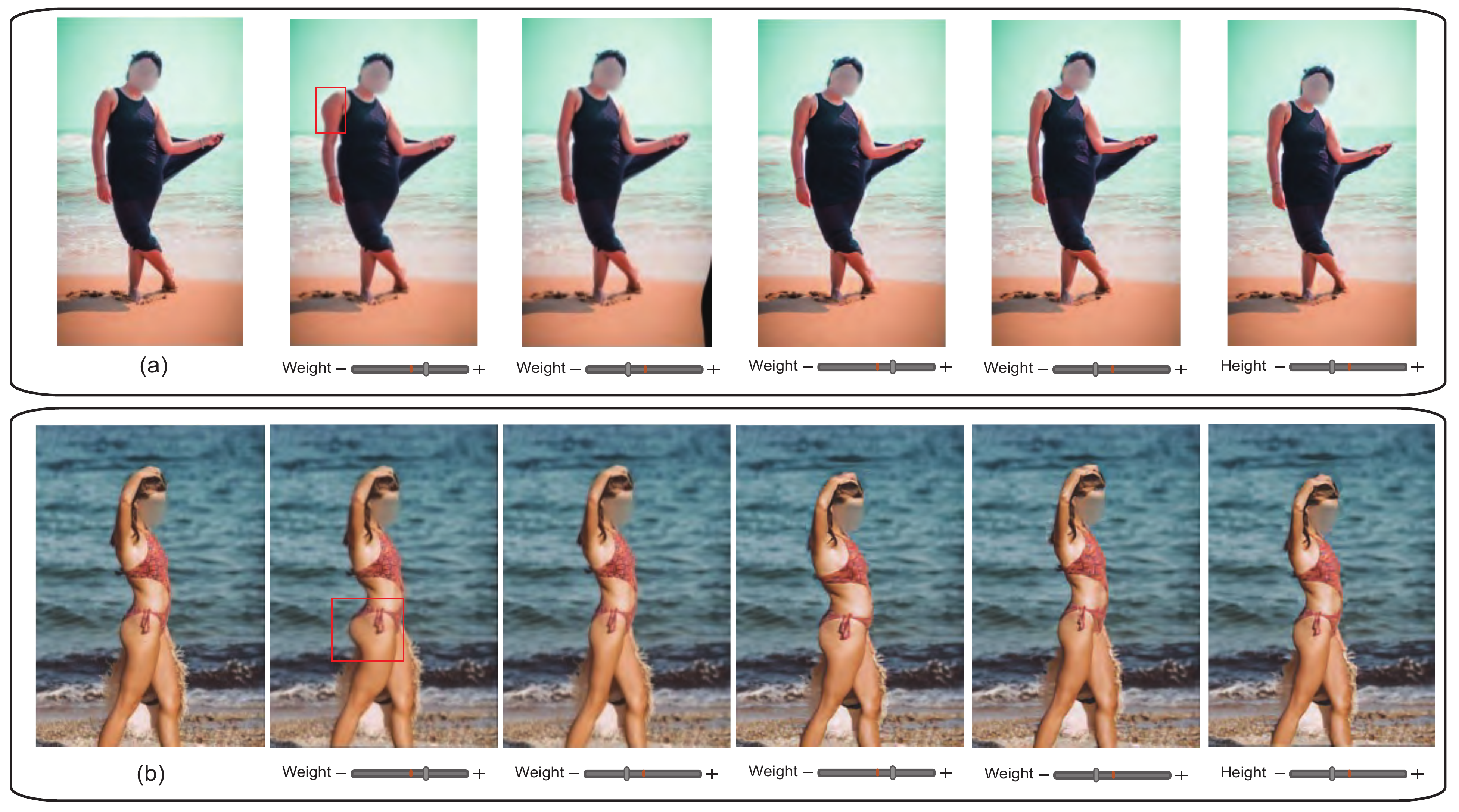}
        \caption[width=18]{Comparison with Ren et al. \cite{ren2022structure} on BR-5K dataset. For each case, from left to right are the original image, the two reshaping results (i.e. weight-gaining and weight-losing) by \cite{ren2022structure}, and three reshaping results (i.e. weight-gaining, weight-losing, the change of height) by our method.} 
	\label{fig:comparasion_sota_br5k}
\end{figure*}

We also compare our method with the state-of-the-art reshaping method~\cite{ren2022structure}. Ren et al.~\cite{ren2022structure} exploit neural networks to generate deformation flow for image warping. A single scalar is used to control the deformation extent. To train their network, they require a dataset of reshaped ground truth produced by artists. However, their method can only amend the weight and is unable to modify the height. Visual comparisons on COCO dataset are shown in Fig. \ref{fig:comparasion_sota}. The results show that our method generates comparable results with \ref{fig:comparasion_sota} (see results in columns 2 and 3) while our method can also change the height (see results in column 4). Moreover, the deformation field generated by their method can introduce artifacts to the image. As highlighted in the red bounding boxes in Fig .\ref{fig:comparasion_sota} (c), undesirable deformations and distortions exist, while our method generates more realistic results. For a fairer comparison, we also present results on their released dataset BR-5K in Fig. \ref{fig:comparasion_sota_br5k}. % For each case, from left to right are the original image, two reshaping results (i.e. weight-gaining and weight-losing) produced by their method, and two corresponding results generated by our method. The last column shows the case of height changing produced by our method. From Fig. \ref{fig:comparasion_sota_br5k}, 
We again observe that Ren et al. \cite{ren2022structure} sometimes generate unnatural reshaping effects, as highlighted in the red bounding box, while our method produces more global consistent results. Moreover, their method%, which controls the reshaping effect using a single scalar, 
is unable to produce height change, while our method generates realistic height-changing effects.

\begin{table}[h]
	\centering
	\begin{tabular}{cc}
		\toprule
		Method & FID $\downarrow$\\
		\midrule
		Liquid Warping \cite{liu2019liquid} & 89.41673 \\
		Ours & \bf80.28321 \\
		\bottomrule
	\end{tabular}
	\caption{Quantitative evaluation of the generated images with the retouched new poses.}
	\label{tab:fid}
\end{table}

\begin{figure}[h]
	\centering
	\includegraphics[width=0.8\linewidth]{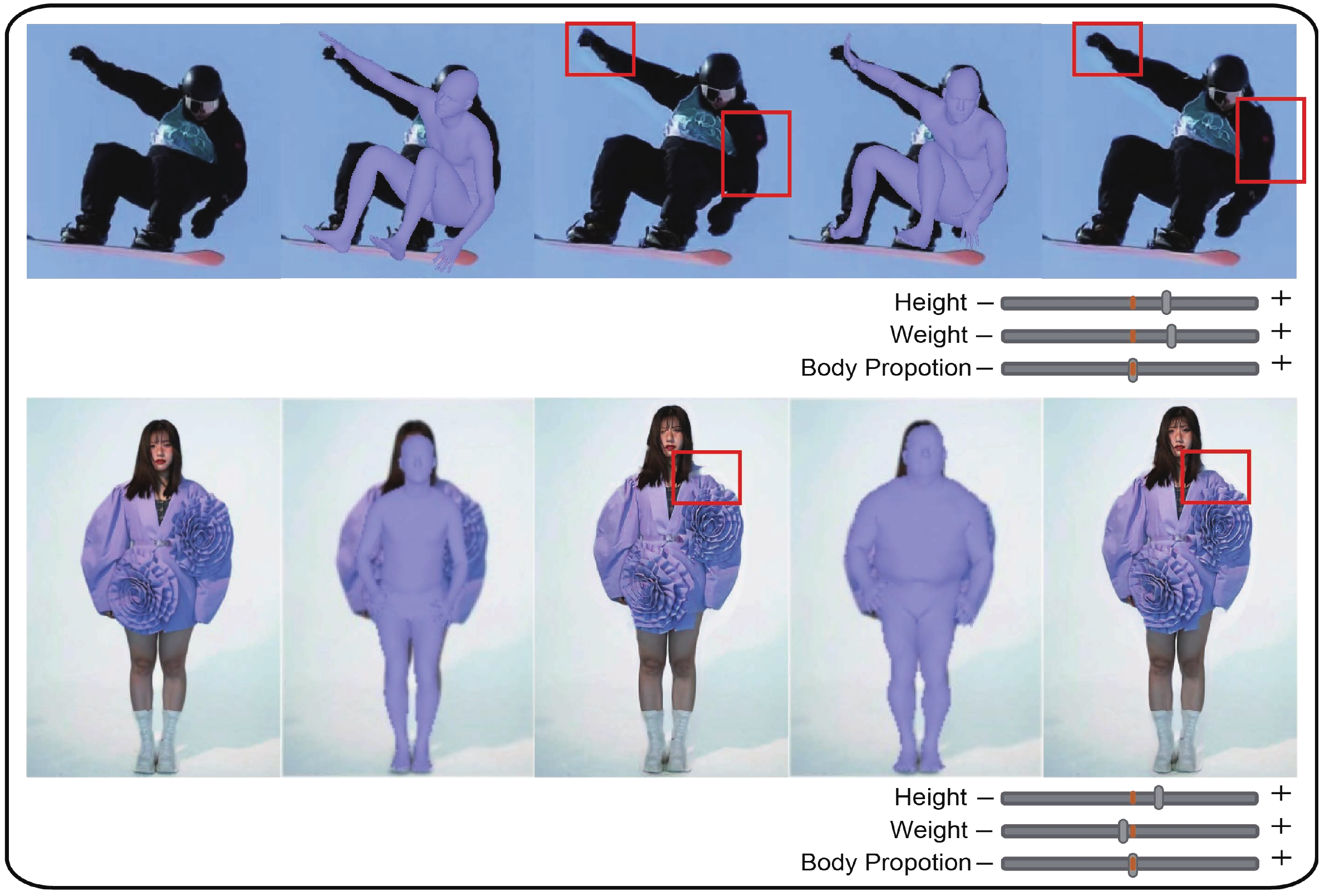}
	\caption[width=18]{Ablation study on the SMPL fitting optimization. 
		The first row shows the fitted SMPL model and synthesized image without 2D keypoints optimization and the corresponding results with our full optimization. The second row shows the comparison on the silhouette optimization.}
	\label{fig:smpl_ablation}
\end{figure}

\subsection{Ablation Study}
We put considerable efforts into the network design to keep it compact without compromising the reshaping performance. To analyze the impact of each component and justify its necessity, we perform two ablation studies. We refer to the network introduced by Fig. 2 in our main paper  as the full model. We obtain the variants by replacing different components with alternatives and train them with the same protocol.

First, we demonstrate the necessity of our optimization during the SMPL model fitting stage. As shown in Fig. \ref{fig:smpl_ablation}, without either of our optimization, there exist feature loss and distortions in the synthesized images. This is because if we fail to fit the SMPL model to the body region, we cannot compute the accurate warping field for reshaping. Specifically, using 2D keypoints for optimization will improve our performance when facing challenging poses, and using the 2D silhouette for optimization will improve our performance when handling fluffy clothes. Note that, although our method can handle fluffy clothes like Fig. \ref{fig:smpl_ablation}, we will also have distortions for cases with very loose skirts or flying hairs.

\begin{figure}[t!]
	\centering
	\includegraphics[width=0.8\linewidth]{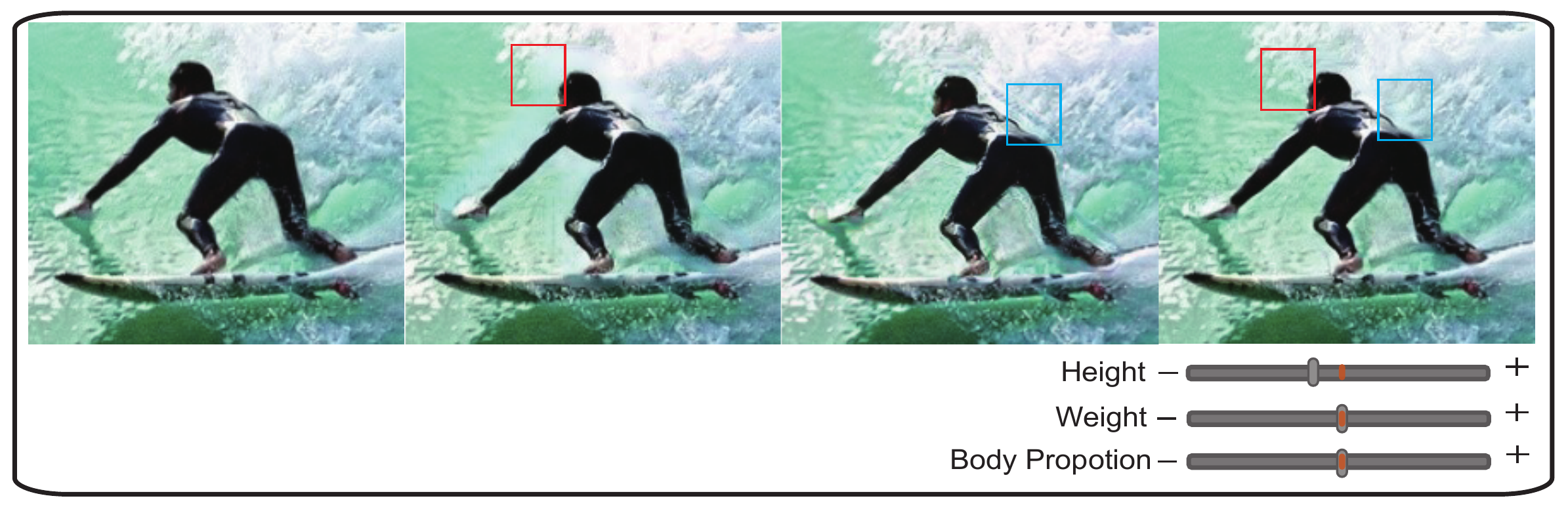}
	\vspace{-0.2cm}
	\caption[width=18]{Ablation study. 
		For each case, from left to right are the original image, the reshaping result generated by \textbf{G-}, the result by \textbf{M+}, and the result by our full model.
		The sliders below each case indicates the attributes and their degrees that have been edited. Red and blue rectangles are used to highlight the areas with artifacts.}
	\label{fig:comparasion_cov}
	\vspace{-0.3cm}
\end{figure}

In reshaping stage, we replace the gated convolution with vanilla convolution for the background branch. We denote the resulting model as variant \textbf{G-}. Fig. \ref{fig:comparasion_cov} shows that the synthesized textures by \textbf{G-} %its synthesized textures 
tend to be blurry and exhibit color discrepancy with their surroundings. In contrast, our full model successfully produces %succeeds to produce 
rich and consistent details (see the area highlighted with a red rectangle).

\begin{figure}[t!]
	\centering
	\includegraphics[width=\linewidth]{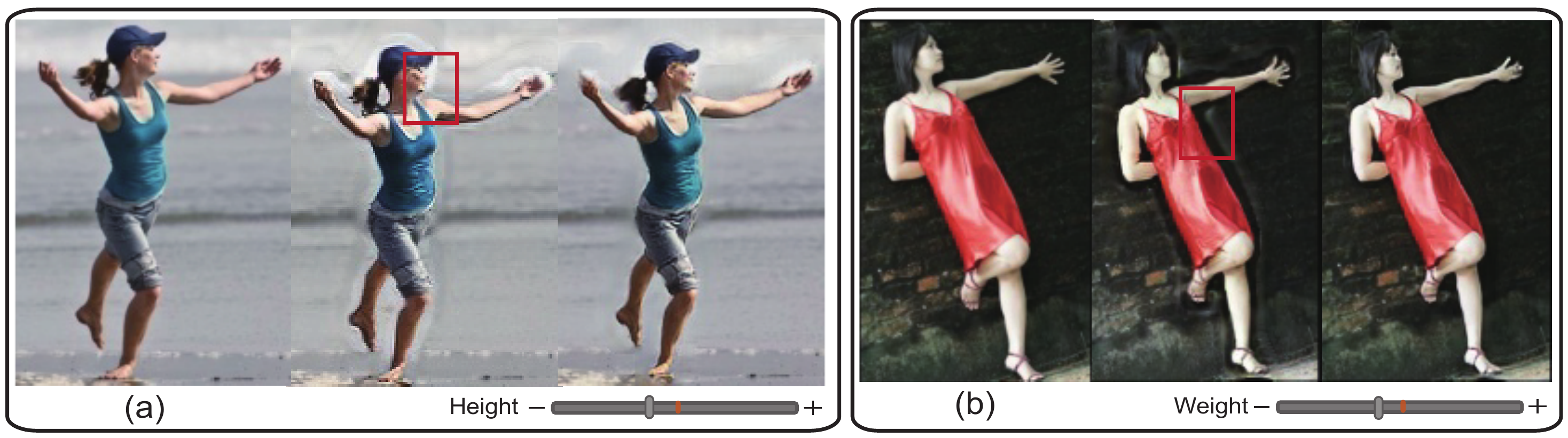}
	\vspace{-0.2cm}
	\caption[width=18]{Ablation study. 
		For each case, from left to right are the original image, the reshaping result generated by \textbf{C-} and the result by our full model. The sliders below each case indicates the attributes and their degrees that have been edited. Red rectangles are used to highlight the areas with artifacts.}
	\label{fig:com_more}
	\vspace{-0.3cm}
\end{figure}

The image reshaping requires an extensive spatial rearrangement on a complex background. The key is how to blend the deformed foreground with the inpainted background without introducing artifacts. Our method exploits the simple addition of the two branches in the feature space. To validate its effectiveness, we compare it with two alternatives. The first one is a popular combine method that merges two feature maps in a mask-guided way \cite{tan2020michigan}. Specifically, we compute a mask for the warped foreground, denoted as $m^{t}$. We then integrate the features as
\begin{equation}
	\label{masked_guided}
	\varphi_{i} = b_{i} + f_{i}^{t} * m^{t},
\end{equation}
where $b_{i}$ and $f_{i}$ are the background and foreground feature maps of the $i$-th layer, respectively, and $*$ denotes the element-wise multiplication. We replace Eq. 4 in our main paper with Eq. \ref{masked_guided} in the full model to obtain the variant \textbf{M+}. Fig. \ref{fig:comparasion_cov} shows that the mask-guided merging tends to introduce artifacts around synthesized humans (see the area highlighted with a blue rectangle). The second way is that we only merge features after the last layer of encoding. Thus the skip-connections are not available in such situation. We denote the resulting model as variant \textbf{C-}. Fig. \ref{fig:com_more} shows that without per-layer merges and skip-connections, our method tends to introduce artifacts around synthesized humans (see areas highlighted with red rectangles).

\begin{figure*}[t!]
	\centering
	\includegraphics[width=\linewidth]{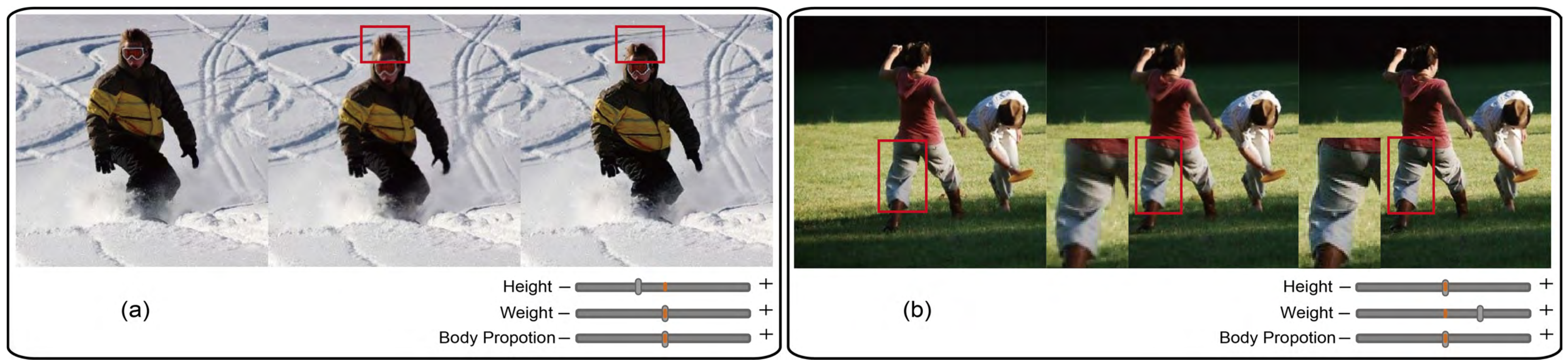}
	\vspace{-0.3cm}
	\caption[width=18]{Comparison with an alternative approach by combining direct warping on foreground with background inpainting. For each case, from left to right are the original image, the reshaping result by the alternative approach and the reshaping result by our method.}
	\label{fig:comparasion_warp}
	\vspace{-0.3cm}
\end{figure*}

We compare our method with an alternative that directly warps the foreground (instead of the feature space) and composes it with the inpainted background. We diffuse the warping field to the whole foreground mask via the warping strategy used in \cite{geng2018warp} and employ the network presented in \cite{yu2019free} for inpainting. Fig. \ref{fig:comparasion_warp} shows that our method synthesizes realistic shapes and details while the direct-warping strategy often brings in distortions (e.g., the hair region in Fig. \ref{fig:comparasion_warp} (a)) or blurriness (Fig. \ref{fig:comparasion_warp} (b)).

\section{Limitations}
\begin{figure}[t!]
	\centering
	\includegraphics[width=\linewidth]{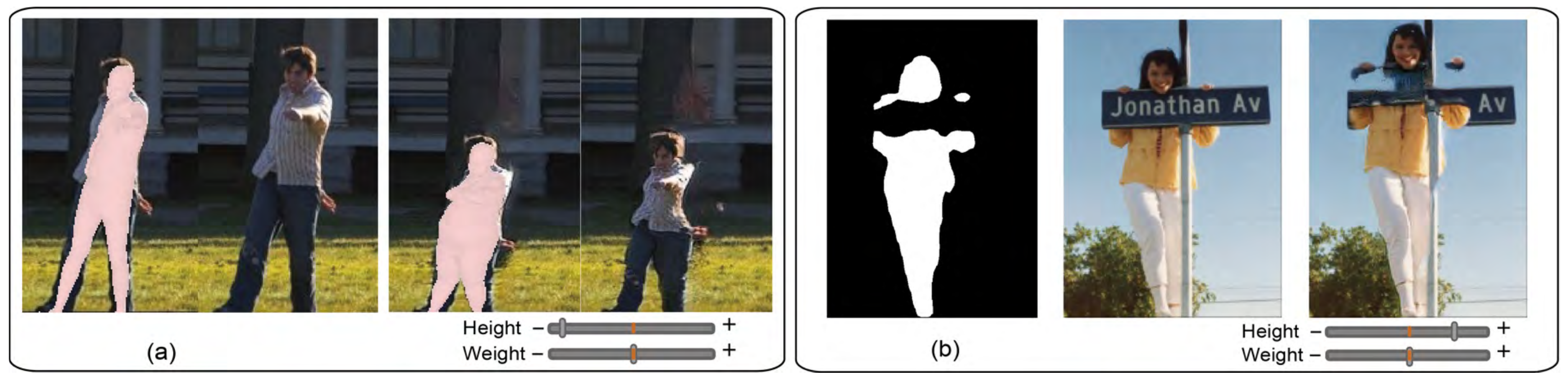}
	\vspace{-0.3cm}
	\caption[width=18]{Failure results generated by our method. (a) Synthesized results under extremely large deformations. (b) Synthesized results when the interacting objects lie outside the foreground mask.}
	\label{fig:limitations}
	\vspace{-0.3cm}
\end{figure}

Our method has a few limitations. Our method might introduce artifacts under extreme deformations (see Fig. \ref{fig:limitations} (a)). Most of the results we presented are within a weight/height change of 20 kg/cm to ensure the deformed SMPL model is valid in 3d space. However, when deformations larger than this range happens, our method may fail. The reasons are two-fold. First, the resculpted 3D model might become implausible, leading to unnatural image warping effects. Second, large deformations on human body can cause undesirable distortions on human faces when the SMPL face model is not fitted well with face areas on 2d images. Third, the inpainting may fail, since such a large occlusion would not appear in the training data. To alleviate such problems, we could learn a manifold that admits valid SMPL parameters, and reject implausible ones. Moreover, we could enhance our network's texture hallucination for large occlusions by improving the data and strategies.

Our method relies on the 3D fitting, whose imperfections may degrade the image reshaping effects. The fitting could fail in cases where severe occlusions {or self-occlusions} exist. We observe that the fitting quality decreases dramatically if more than half of the human body is dis-occluded. Manual corrections may help in such situations. On the other hand, the automatic fitting of SMPL model often fails at fine-grained regions, such as hands, which may lead to undesirable synthesized results at those regions (Fig. \ref{fig:deepfashion_results}, \ref{fig:in_the_wild}, \ref{fig:internet_cases}). A more fine-grained parametric model \cite{pavlakos2019expressive} might help.

Our method deals well with parts tightly coupled with the body, like cloth, hat, or hair, since Mask R-CNN \cite{he2017mask} usually segments the foreground with these areas included. However, if the interacting objects with people are not covered by the segmentation mask, they remain where they were during the synthesis, thus decreasing the realism of the generated results (see Fig. \ref{fig:limitations} (b)). To appropriately represent the interactions between objects, we could introduce the scene graphs into the generation process.

For generating high resolution results, our method needs high resolution datasets. However, training with high resolution data is very time-consuming and requires large GPU memories. Training with multi-scale generator and discriminator like~\cite{wang2018high} might help. Also, as to the human face, adding constraints to fix face features may help us avoid face changing.

Our method cannot handle the reshaping of multiple people simultaneously. The system requires the localization of the reshaped person via a bounding box to enable the sequential editing of each individual in multi-person images. 

\end{appendix}

\end{document}